\documentclass{article}

\PassOptionsToPackage{numbers,compress}{natbib}

\usepackage{multirow}
\usepackage[preprint]{neurips_2023}
\usepackage{graphicx,subfigure,xspace,verbatim,comment}
\usepackage{tabularx}
\newcolumntype{Y}{>{\centering\arraybackslash}X}

\usepackage[utf8]{inputenc} 
\usepackage[T1]{fontenc}    
\usepackage{hyperref}       
\usepackage{url}            
\usepackage{booktabs}       
\usepackage{amsfonts}       
\usepackage{nicefrac}       
\usepackage{microtype}      
\usepackage{xcolor}         
\newcommand{\system}{\textsc{Saturn}}

\title{Saturn: Efficient Multi-Large-Model Deep Learning}

\author{%
  Kabir Nagrecha \\
  \texttt{knagrech@ucsd.edu} \\
  \And
  Arun Kumar \\
  \texttt{arunkk@eng.ucsd.edu} \\
  }

\begin{document}

\maketitle

\begin{abstract}
  In this paper, we propose \system, a new data system to improve the efficiency of \textit{multi-large-model training} (e.g. during model selection/hyperparameter optimization). We first identify three key interconnected systems challenges for users building large models in this setting --- parallelism technique selection, distribution of GPUs over jobs, and scheduling. We then formalize these as a joint problem, and build a new system architecture to tackle these challenges simultaneously. Our evaluations show that our joint-optimization approach yields 39-49\% lower model selection runtimes than typical current DL practice.
\end{abstract}

\vspace{-5mm}

\section{Introduction}
Fine-tuning large-scale deep learning (DL) models has become a common practice in a variety of domains. It is now common for a developer to download an open-source model spanning tens of billions of parameters from a model hub (e.g. HuggingFace~\cite{huggingface2019}) and fine-tune it on their application-specific data~\cite{wang2021gradient}. Unfortunately, models of these sizes tend to impose considerable time \& cost burdens, often making them impractical for users in smaller companies \& the domain sciences~\cite{kumar2021cerebro,nakandala2020incremental}. 

We identify a few key systems-oriented headaches for users seeking to fine-tune large models: (1) \textit{GPU memory} remains a bottleneck. Large-memory GPUs are expensive, and even public cloud vendors still ration them. (2) \textit{Multi-GPU parallelism} is needed, but dozens of parallelism techniques exist~\cite{swapadvisor2021,gpipe2018,torchfsdp2021,zero2019,megatron2019,alpa2022,hotline2022,hydra2021,mpms2021,techreport,researchExam,terapipe2021,fedus2021switch}, and understanding the performance behaviors of these complex approaches can be difficult for DL users. (3) \textit{Model selection}, i.e. tuning hyper-parameters, model layers, etc., only amplifies the computational load by producing multiple models to train.

These challenges result in a three-part technical problem for practitioners.
First, \textbf{resource allocation}~\cite{nagrecha2023intune}. Given a cluster of GPUs, how should they be distributed across jobs? Second, \textbf{parallelism selection}. For each model in a multi-job, which technique should be applied? Third, \textbf{scheduling}. What job ordering will produce the lowest makespan?

\textbf{\textit{These three questions are deeply intertwined.}} GPU allocations affect the optimal parallelism selection for each job, and constrain schedule orderings. Per-job parallelism selections affect runtimes, thus impacting scheduling \& allocation decisions. In this paper, we describe a new approach to tackling this intertwined problem, and demonstrate how taking a novel ``joint optimization'' approach can significantly improve performance.

\section{System Architecture \& Approach}

\begin{figure*}[t]
\includegraphics[width=\textwidth]{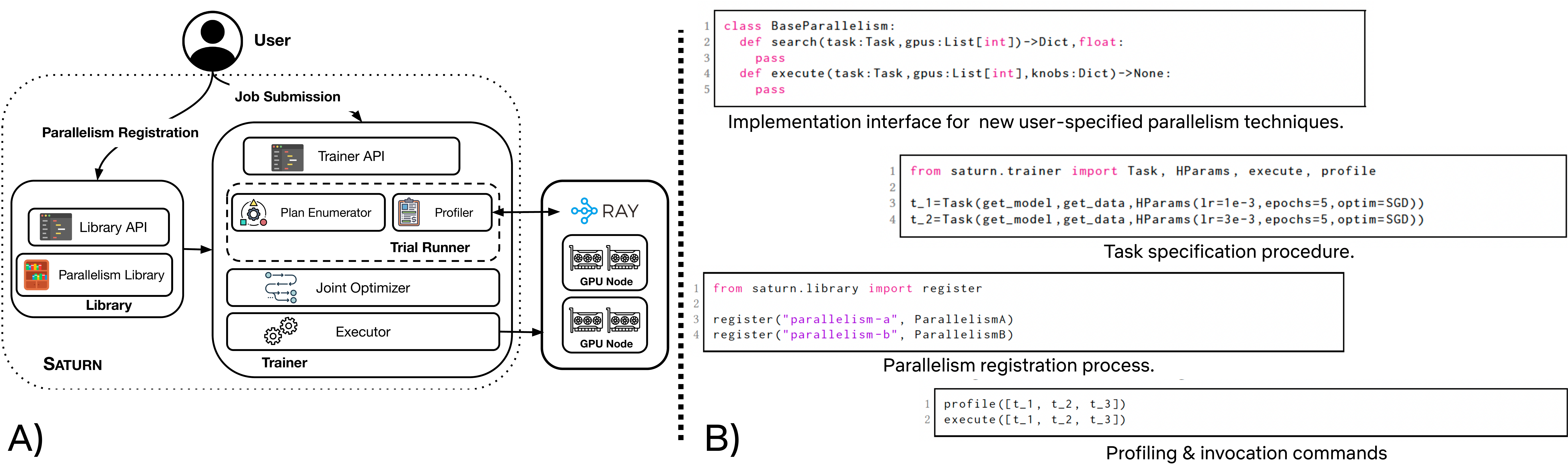}
\vspace{-7mm}
\caption{(A) System architecture of~\system~and the interactions between the components. (B) Overview of key calls in \system's API.}
\label{fig:sys_architecture}
\vspace{-5mm}
\end{figure*}

\system~has 3 main modules --- the Parallelism Library, the Trial Runner, \& the Solver. Figure~\ref{fig:sys_architecture}(A) illustrates the overall design. 

The Parallelism Library allows users to register and apply various model parallelization techniques with minimal effort. Techniques can be added to the Library by implementing a simple two-function interface, as shown in Figure~\ref{fig:sys_architecture}(B). They can then be registered to the library to be reused in different execution sessions (even across different cluster users).

The Trial Runner is our empirical approach to profiling \& evaluating user-provided black-boxes (models, parallelisms). Given a set of models, the Trial Runner profiles each model under each possible parallelism under each possible GPU count it could be assigned. Since profiling only requires processing one or two mini-batches, this profiling time tends to be negligible in the context of a larger job. The information from the Trial Runner is critical for our next component, the Solver.

The Solver uses the empirical estimates from the Trial Runner to formulate our joint optimization problem --- parallelism selection, resource distribution, and scheduling --- as a mixed-integer linear program (MILP). We apply the popular MILP solver Gurobi~\cite{gurobi} to produce solutions. The output is used to determine: (1) which parallelism technique from the Library each model should use, (2) how many GPUs each model should get, and (3) in what order \& when the models should be submitted to the cluster. We augment the solver with an ``introspection'' mechanism, adapted from prior art~\cite{xiao2018gandiva,xiao2020antman}. As models are trained, remaining runtimes per-model will change and shift the workload. So, we \textit{re-run} the solver on fixed intervals. When a new plan is produced,  executing jobs are checkpointed and re-launched under the new scheme.

\section{Experiments}
We evaluate \system's performance against four baselines from standard practice \& prior art. The first, ``Current Practice'', involves allocating all GPUs per node to one job at a time, with models running in sequence. Task parallelism is applied across nodes. The second baseline, ``Random'', randomizes allocations, parallelisms, and schedule orderings. The third baseline ''Optimus''~\cite{peng2018optimus}, greedily assigns GPUs one-at-a-time based on the estimated marginal runtime improvement. The fourth baseline, Optimus-Dynamic, augments Optimus with our introspection mechanism. Since all approaches optimize on a higher orchestration level, they do not affect correctness/accuracy. We register 4 parallelism techniques in the Library for \system: FSDP \& DDP from PyTorch Distributed, GPipe, and model offloading from FairScale. Table~\ref{tb:details} provides the details of our workloads. 

\begin{table*}
\centering
\caption{Experimental details.}
\vspace{1mm}
\begin{tabularx}{\textwidth}{ c c  c  c  c  c } 
\toprule
Hardware & Epochs & Learning Rates & Batch Sizes & Models & Datasets \\
\midrule
\multirow{2}{*}{p4d.24xlarge} & \multirow{2}{*}{10} & \multirow{2}{*}{1e-5/1e-4/1e-3} & 16/32 & GPT-2/GPT-J & WikiText-2~\cite{wikitext-2}  \\ 
\cmidrule(lr{1em}){4-6}
& & & 64/128 & ViT-G/ResNet-200 & ImageNet~\cite{imagenet}
\label{tb:details}
\end{tabularx}
\vspace{-7mm}
\end{table*}

Table~\ref{tb:experiments} demonstrates our results from both workloads. We evaluate first on a single 8-GPU node, then on two such nodes. In both cases, \system~significantly outperforms the baselines. We see speedups of 1.64-1.96X versus the Current Practice baseline, with training time reductions of 39-48\%. We find that the allocations \system~chooses are often unintuitive to a human  (e.g. giving 5 GPUs to one model and applying GPipe, but 3 GPUs to another with FSDP. But the overall result is lower costs \& runtimes. These results indicate that using \system~for multi-large-model training may help encourage adoption of large models by time- \& and cost- constrained users.

\begin{table*}[b!]
\centering
\caption{Runtimes (hours) from our experiments. Numbers are reported as (1-node/2-node).}
\vspace{1mm}
\begin{tabular}{c c c c c c} 
& Current Practice & Random & Optimus & Optimus-Dynamic & \system \\
\toprule
\textbf{WikiText} & 28.39/14.57 & 41.45/21.76 & 34.9/16.62 & 24.87/13.62 & 17.24/8.23  \\ 
\textbf{ImageNet} & 19.05/10.15 & 28.34/14.44 & 19.44/10.19 & 17.31/8.32 & 11.31/5.16  \\ 
 \label{tb:experiments}
\end{tabular}
 \vspace{-10mm}
\end{table*}

\clearpage

\bibliographystyle{acm}
\bibliography{main}

\end{document}